\documentclass[final]{cvpr}
\usepackage{times}
\usepackage{epsfig}
\usepackage{graphicx}
\usepackage{amsmath}
\usepackage{amssymb}
\usepackage{booktabs}
\usepackage[accsupp]{axessibility}

\usepackage{graphics} % for pdf, bitmapped graphics files
\usepackage{graphicx}
\usepackage{grffile} %file name with multi-dots
\usepackage{amsmath} % assumes amsmath package installed
\usepackage{amssymb}  % assumes amsmath package installed
\usepackage{color}
\usepackage{bm}
\usepackage{url}

\usepackage[noend]{algpseudocode}
\usepackage{algorithm}
\usepackage{makecell}
\usepackage[square, comma,numbers,sort&compress]{natbib}
\usepackage{multirow}
\usepackage{enumitem}
\usepackage[table,dvipsnames]{xcolor}
\usepackage{xcolor}
\usepackage[pagebackref=true,breaklinks=true,colorlinks,bookmarks=false]{hyperref}

\hypersetup{
linkcolor=BrickRed
,citecolor=Green
,filecolor=Mulberry
,urlcolor=NavyBlue
,menucolor=BrickRed
,runcolor=Mulberry
,linkbordercolor=BrickRed
,citebordercolor=Green
,filebordercolor=Mulberry
,urlbordercolor=NavyBlue
,menubordercolor=BrickRed
,runbordercolor=Mulberry
}

\newcommand{\siyuan}[1]{{\textcolor{black}{#1}}}
\newcommand{\anbang}[1]{{\textcolor{black}{#1}}}
\def\confYear{CVPR}

\begin{document}

\def\cvprPaperID{11115}
%%%%%%%%% TITLE
\title{Self-supervised Spatial Reasoning on Multi-View Line Drawings}

\author{
 {Siyuan Xiang\thanks{Equal contribution.} \ \footnotemark[3]{}}
 \and
 {Anbang Yang\footnotemark[1]{} \ \footnotemark[3]{}}
 \and
 {Yanfei Xue\footnotemark[3]{}}
 \and
 {Yaoqing Yang\footnotemark[4]{}}
 \and
 {Chen Feng\thanks{The corresponding author is Chen Feng {\tt\small cfeng@nyu.edu}.} \ \footnotemark[3]{}}
 \and
 \normalsize{\textsuperscript{$\ddagger$}New York University Tandon School of Engineering \quad \textsuperscript{$\mathsection$}University of California, Berkeley}
 \\ \url{https://ai4ce.github.io/Self-Supervised-SPARE3D/}
}

\maketitle

%%%%%%%%% ABSTRACT
\begin{abstract}
Spatial reasoning on multi-view line drawings by state-of-the-art supervised deep networks is recently shown with puzzling low performances on the SPARE3D dataset~\cite{han2020spare3d}. 
% To study the reason behind the low performance and to further our understandings of these tasks, we design controlled experiments on both input data and network designs.
Based on the fact that self-supervised learning is helpful when a large number of data are available, we propose two self-supervised learning approaches to improve the baseline performance for \textit{view consistency reasoning} and \textit{camera pose reasoning} tasks on the SPARE3D dataset.
For the first task, we use a self-supervised binary classification network to contrast the line drawing differences between various views of any two similar 3D objects, enabling the trained networks to effectively learn detail-sensitive yet view-invariant line drawing representations of 3D objects. 
For the second type of task, we propose a self-supervised multi-class classification framework to train a model to select the correct corresponding view from which a line drawing is rendered. Our method is even helpful for the downstream tasks with unseen camera poses.
Experiments show that our method could significantly increase the baseline performance in SPARE3D, while some popular self-supervised learning methods cannot.

\vspace{-6mm}
\end{abstract}

%%%%%%%%% BODY TEXT

\section{Introduction}\label{sec:intro}
Human visual reasoning, especially spatial reasoning, has been widely studied from psychological and educational perspectives~\cite{kell2013creativity,hsi1997role}. Researches show that trained humans can achieve good performance on spatial reasoning tasks~\cite{ramful2017measurement} because they can solve these tasks using spatial memory, logic, and imagination. However, the spatial reasoning of deep networks is yet to be explored and improved. In other visual learning tasks such as image classification, object detection, and segmentation, state-of-the-art deep networks have shown their superior performance to humans by memorizing indicative visual patterns from enormous image instances for prediction. However, it seems difficult for deep networks to reason using the same mechanism about the spatial information such as the view consistency and camera poses from 2D images~\cite{han2020spare3d}.

\begin{figure}[t]
	\centering
	\includegraphics[width=0.91\columnwidth]{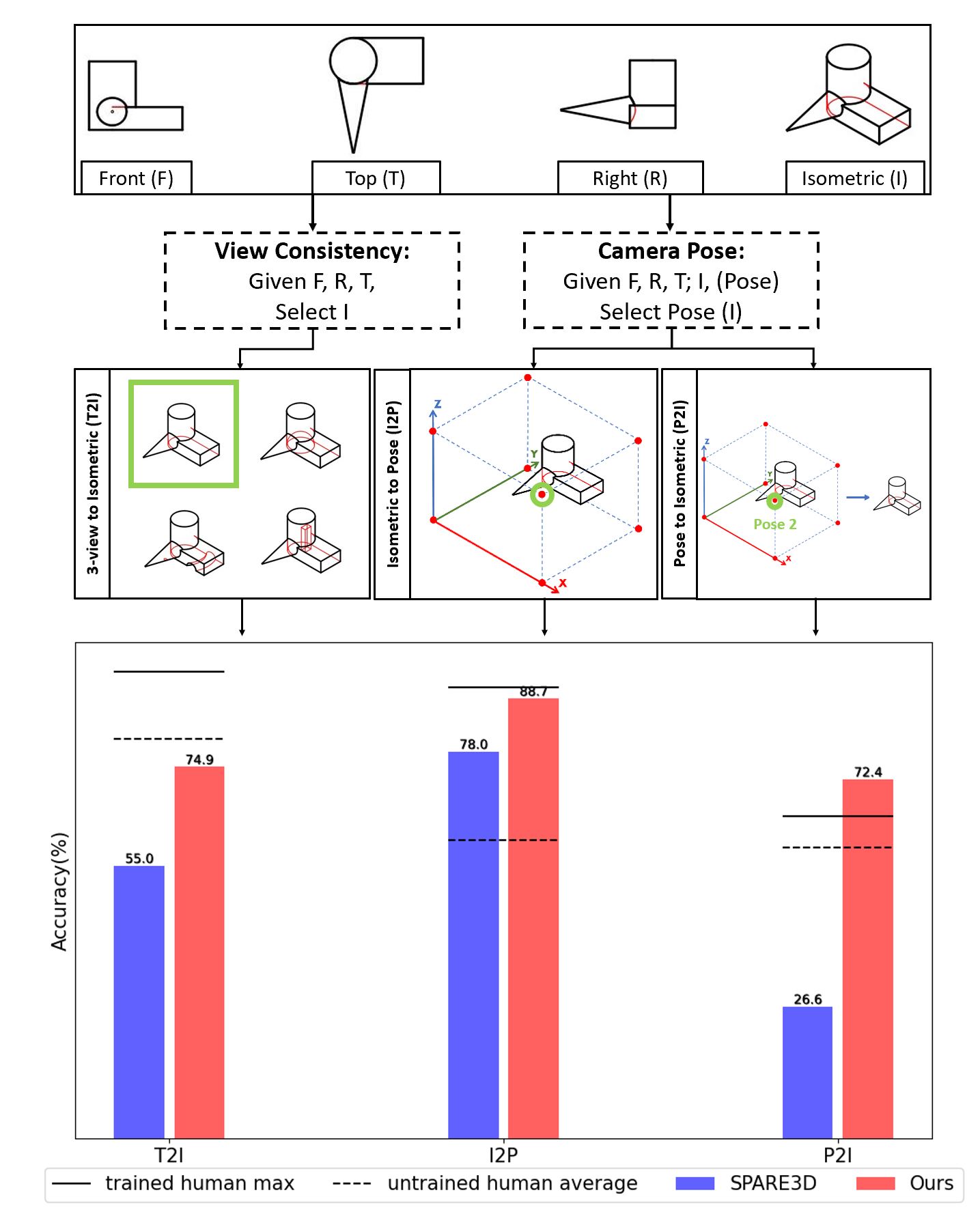}
	
	\caption{We significantly improve SPARE3D baselines using self-supervised learning approaches. Note that in this figure, all the networks are trained using the same amount of data (data generated from 5,000 CAD models) as used on the SPARE3D dataset.
	}
	\label{fig:compare}
	\vspace{-5mm}
\end{figure}

To our best knowledge, we are the first to investigate spatial reasoning tasks in the SPARE3D\footnote{Note that we use the latest dataset and benchmark results updated by the SPARE3D authors after CVPR 2020.} dataset~\cite{han2020spare3d}, that provides several challenging spatial reasoning tasks in line drawings (Figure~\ref{fig:compare}). The dataset is unique in that: (1) it uses line drawings as inputs, (2) it is a non-categorical dataset, meaning there is no class label information for each object, (3) it defines spatial reasoning instead of recognition tasks.
Specifically, we focus on the view-consistency reasoning task (the Three-view to Isometric or \textit{T2I}) and the two camera-pose reasoning tasks (the Isometric to Pose or \textit{I2P}, and the Pose to Isometric or \textit{P2I}). The task examples are illustrated in Figure~\ref{fig:compare} and for more details of the task settings, we refer the readers to the SPARE3D paper~\cite{han2020spare3d}.
% The view-consistency reasoning task requires a deep network to select the correct isometric view drawing out of four visually similar candidates that are consistent with the three-view (front, right, and top views) drawings given in the question.
% For the camera-pose reasoning tasks, given three-view drawings, the network is asked to select the camera pose \siyuan{corresponding to} a given isometric drawing or \siyuan{contrarily,} select the correct isometric drawing \siyuan{corresponding to} a given viewing pose.

% \siyuan{We believe solving these tasks could bring three general benefits or inspiration for the computer vision community. First, it could benefit the \textit{object level} 3D model reasoning. Second, since the networks proposed in this paper are not specially designed for line drawings, we believe our methods are not limited to line drawings, but could also be used for the tasks using RGB images.
% Lastly, the design of the camera pose reasoning tasks could be related to reasoning tasks that study object or camera rotation and manipulation, e.g.~\cite{kendall2015posenet}. With the ability to solve view consistency task and camera pose reasoning task, one scenario could be, given a single view image captured by a sensor, a robot can first understand it is seeing which 3D model in the database, then determine the camera pose itself is looking from.}

To improve the DNN's performance, our first effort is to explore the supervised learning network's ability in three aspects: (1) the quantity of data used for learning, (2) the network's capacity (width and depth), and (3) the network's structure. However, we find changing these factors could not bring significant improvement (all these results are shown in the supplementary), while it might cost more time and computational resources to generate data for supervised learning.

Compared with supervised learning, self-supervised learning is helpful when a large number of unlabeled data are available.
Moreover, in many visual learning tasks, self-supervised learning as a pre-training task can learn better visual representations that can be further fine-tuned via supervision for downstream tasks, achieving similar or even better results than supervised learning~\cite{chen2020big}. Based on these findings, we design two self-supervised learning methods for both view consistency reasoning task and camera pose reasoning task on the SPARE3D dataset.

\begin{figure*}[t]
	\centering
	\includegraphics[width=0.96\linewidth]{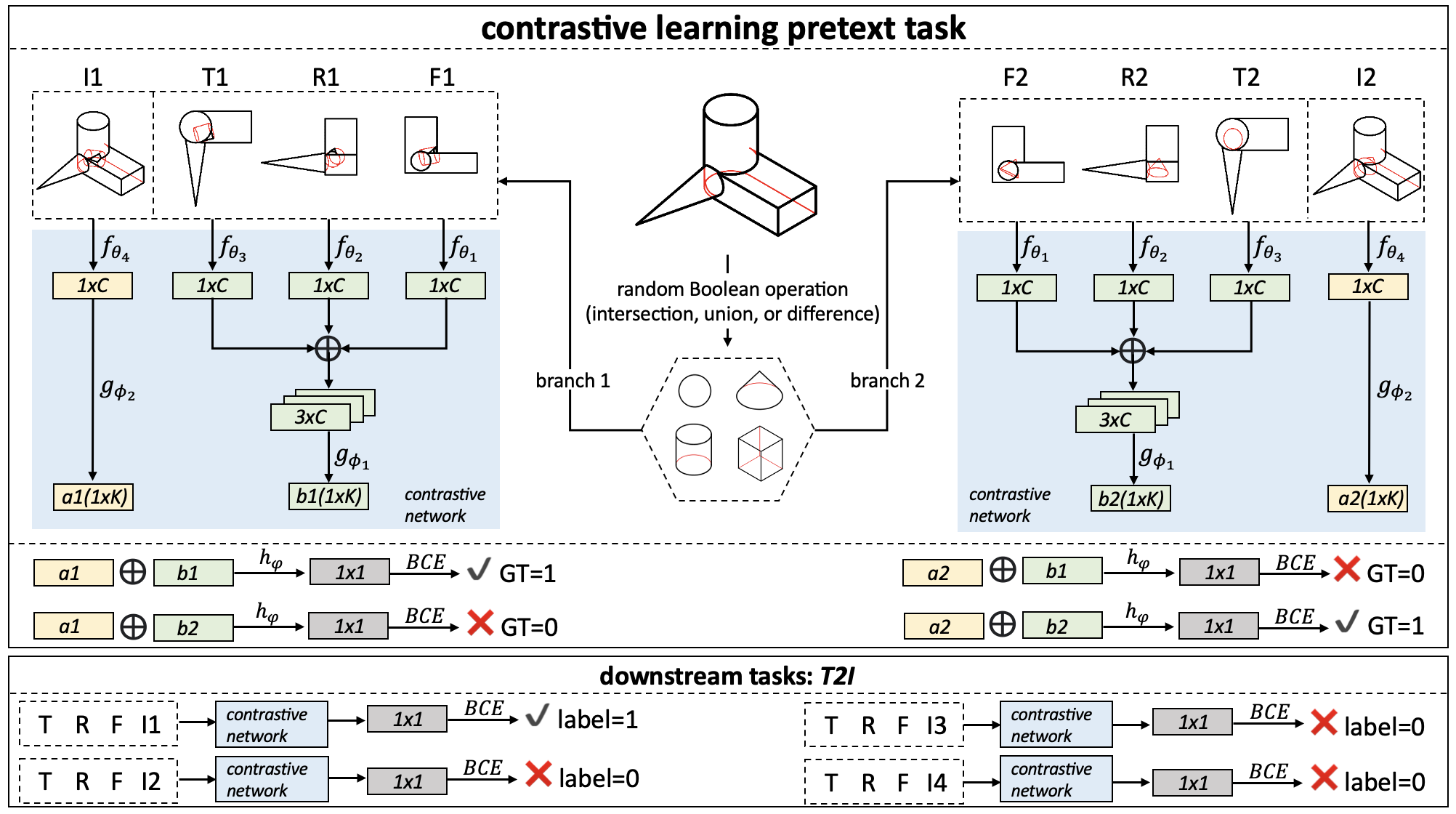}
	\caption{\textbf{Contrastive spatial reasoning architecture for task \textit{T2I}.} We train a contrastive learning network, then use the learned representations to the downstream task \textit{T2I}.
% 	For the contrastive learning network,} a CAD model is used to generate two CAD models with different Boolean operations and primitives like sphere, cone, cylinder, or cube. Then two sets of multi-view line drawings for the two new models are rendered respectively, in branch 1 and 2. Each set is composed of three-view line drawings including 
	Front (F), Right (R), Top (T) represent line drawings and an isometric (I) line drawing. 
    $a_{1}, b_{1}, a_{2}, b_{2}$ are the encoded feature vectors; $C$, $K$ are the dimension of the latent vectors. The $f_{\theta_1}$, $f_{\theta_2}$, $f_{\theta_3}$, and $f_{\theta_4}$ are CNN networks; $g_{\phi_1}$, $g_{\phi_2}$, and $h_\psi$ are MLP networks. $\bigoplus$ is a concatenation operation. BCE represents binary cross-entropy loss. }
%   \siyuan{For solving task \textit{T2I}, we send three view drawings with each candidate isometric view image to the pre-trained network, obtaining the probability for each candidate answer and compare it with ground truth.} GT is used for determining whether two latent vectors are similar, while label is used for }
	\label{fig_contrastive_network}
% 	\vspace{-5mm}
\end{figure*}

 For the view consistency reasoning task, we design a contrastive spatial reasoning method to tackle the challenges in SPARE3D tasks (Figure~\ref{fig_contrastive_network}). This is necessary because most of the existing contrastive learning methods do not explicitly consider the relationship between different views, nor do they force deep networks to focus on detailed differences between images. We demonstrate both qualitative and quantitative results that show \textit{our method helps deep networks capture these small details while being view invariant}, which is crucial for the view consistency task. 

For the camera pose reasoning task, we propose a self-supervised learning network for improving the deep network's ability to find the correlation between the pose and the image rendered from the pose (see Figure~\ref{fig_self-supervised}).
Our experiment results show that our self-supervised learning network is not only helpful for the downstream tasks with seen views but also helpful for the tasks with unseen views. This shows a potential to even improve pose estimation by transferring the learned representations from a camera pose reasoning pretext task to related downstream tasks.

% Note that, all our experiments show that our self-supervised learning methods outperform the supervised learning methods when using the same amount of data for training.

In sum, our major contributions are:
\begin{itemize}[nosep,nolistsep]
    \item A novel contrastive learning method by self-supervised binary classifications, which enables deep networks to effectively learn detail-sensitive yet view-invariant multi-view line drawing representations for view consistency reasoning task;
    
    \item A self-supervised multi-class classification method to learn pose-aware representations from multi-view line drawing images;
    % \item A self-supervised multi-class classification method to help deep networks find the corresponding relationship between the pose and the image rendered from that pose; 
    % \siyuan{Even though the camera poses in the downstream tasks might not be seen in the self-supervised training stage};

    % \item Extensive controlled experiments to improve our empirical understandings of SPARE3D tasks, which further help us improve network design for these tasks;

    \item Significantly improved spatial reasoning performance of deep networks on line drawings based on the above, some of which surpass human performance.
    \vspace{-3mm}
\end{itemize}

% So ``why is the baseline performance on SPARE3D low''? This important question is raised in~\cite{han2020spare3d} with three conjectures: 1) the dataset is non-categorical, 2) the images are line drawings, and 3) spatial reasoning is not image retrieval. 
% The first two conjectures inspire us to first understand the task characteristics, and study the factors that affect the network's performance on those tasks, which leads to our quantitative investigations about a network's ability on the SPARE3D tasks using controlled experiments.

% The last conjecture is related to the intuitive difference between reasoning and other discriminative learning tasks. \citet{bengio2017consciousness} pointed out that reasoning can be considered as \emph{Kahneman’s system 2} problem~\cite{kahneman2011thinking}, which ``requires a sequence of conscious steps'', and is slower, more deliberative, and more logical compared to system 1 problems that are ``fast, instinctive and emotional''.
% This inspires us to rethink the SPARE3D baselines' training mechanism, which is fully supervised. Although there seems to have no consensus yet on how to extend deep networks from system 1 to system 2, we find self-supervised learning, especially contrastive learning, a promising mechanism to improve the baseline SPARE3D performances, due to its similarity to how humans intuitively learn to solve SPARE3D tasks. 
\section{Related Work}\label{sec:related}
\vspace{-2mm}

Self-supervised learning has achieved great success due to the performance improvement on many visual learning tasks~\cite{godard2019digging, misra2020self, kim2020spatially, wang2020self, zhai2019s4l, kolesnikov2019revisiting, you2022class, jing2020self, goyal2019scaling, liu2021auto}. Successful self-supervised learning frameworks use different tricks to create artificial labels based solely on the inputs.
One way is to use spatial information or spatial relationships between image patches in a single image. For example, \citet{gidaris2018unsupervised} designed the pretext task by asking the network to predict image rotations. Another way is to ask the network to recover the positions of shuffled image patches~\cite{kim2018learning, noroozi2016unsupervised, wei2019iterative}, or predict the relative position~\cite{doersch2015unsupervised}. In addition, the color signal contained in an RGB image could also be used.
By recovering the color for grayscale images generated from RGB images, networks can learn the semantic information of each pixel~\cite{zhang2016colorful, larsson2016learning}. Despite the success of the spatial reasoning tasks in the SPARE3D dataset, these methods are ineffective since they only use visual information from a single image. Our method aims to consider the common information between different images and, thus, is more suitable to solve those reasoning problems.

One way of grouping various self-supervised learning methods is to divide them into generative vs. contrastive ones~\cite{liu2020self}. On the one hand, generative ones learn visual representations via pixel-wise loss computation and thus forcing a network to focus on pixel-based details; on the other hand, contrastive ones learn visual representations by contrasting the positive and negative pairs~\cite{anand2020contrastive, you2021momentum, you2021simcvd, you2021self}.
Many researchers explored contrastive learning by comparing shared information between multiple positive or negative image pairs. These methods often attempt to minimize the distance of the features extracted from the same data source and maximize the distance between features from different data sources in the feature space. 
\citet{hjelm2018learning} propose Deep InfoMax based on the idea that the global feature extracted from an image should be similar to the same image's local feature and should be different from a different image's local feature. Based on this method, \citet{bachman2019learning} further use image features extracted from different layers to compose more negative or positive image pairs.
SimCLR~\cite{chen2020simple} differs from the previous two methods in that it only considers the global features of the augmented image pairs to compose positive and negative pairs. SimCLRv2~\cite{chen2020big} make further improvements on Imagenet~\cite{deng2009imagenet} by conducting contrastive learning with a large network, fine-tuning using labeled data, and finally distilling the network to a smaller network. Compared to SimCLR, MoCo~\cite{he2020momentum} stores all generated samples to a dictionary and uses them as negative pairs instead of generating negative pairs in each step. These tactics could help reduce the batch size requirement while still achieving good performance. Differently, SwaV~\cite{caron2020unsupervised} trains two networks that can interact and learn from each other, with one network's input being the augmented pair of another network's input.
Besides, researchers also theoretically analyze why these contrastive learning methods work well~\cite{arora2019theoretical, tosh2020contrastive, lee2020predicting, tian2020understanding}.

However, \textit{it is difficult to apply the aforementioned contrastive learning methods directly to tasks which requires consideration of the relationships between multi-view images}. Contrastive multiview coding~\cite{tian2019contrastive} has a misleading name in our context because that ``multiview'' in fact, means different input representations instead of views from different camera poses. Therefore it is not suitable for SPARE3D tasks. 
\citet{kim2020few} propose a method to solve a few-shot visual reasoning problem on RAVEN dataset~\cite{zhang2019raven}, which is perhaps more relevant to us due to the use of contrastive learning in visual reasoning. Yet because it is designed for analogical instead of spatial reasoning, it is not directly applicable to SPARE3D tasks either.
% \textbf{Controlled experiments on deep learning}.
% There is a branch of research that investigates how data and networks affect deep visual learning performance via controlled experiments. \citet{funke2020notorious} found that using ImageNet pre-trained network parameters for initialization could significantly improve performances on other visual tasks that seem to have a big domain gap with ImageNet, which inspires us to check whether the same approach is effective for SPARE3D tasks because of the similar domain gap.
% Moreover, we study the network's capacity and structure for SPARE3D tasks, inspired by~\cite{tan2019efficientnet, huang2019gpipe, he2016deep, zagoruyko2016wide}. 

% \input{parts/3-scientific_experiment}
\section{Self-supervised Spatial Reasoning}\label{sec:method}

% According to the previous investigation on network architecture, we find that using ImageNet pre-trained parameters to initialize the backbone network will affect the accuracy of all reasoning tasks. 
\siyuan{In this section, we explore a self-supervised learning method to learn the representations for all three tasks in SPARE3D:  a contrastive spatial reasoning network for view consistency reasoning task \textit{T2I}, a self-supervised learning network for camera pose reasoning task \textit{I2P} and \textit{P2I}. For the camera pose reasoning task, we extend the downstream task settings to show our method is helpful when the camera poses in the downstream task are not seen in the pretext task.}

\siyuan{\subsection{Contrastive learning network for task \textit{T2I}}}
% Based on our investigation of Hypothesis 1 (\siyuan{see supplementary}), \textit{T2I} is a \textit{detail-hurt task}, which means the more complex the models are, the poorer performance the network will have. Therefore, we believe \textit{the key to answering a \textit{T2I} question is to find an effective way to capture the tiny difference among all the candidate isometric line drawings}, which will help correctly select the isometric drawings that is consistent with the three-view line drawings in the question. 
According to the discussion in the Introduction~\ref{sec:intro}, we believe contrastive learning can help the network learn better line drawing representations. Because it is not difficult to obtain a large number of unlabeled CAD models, we design our specialized contrastive spatial reasoning method as illustrated in Figure~\ref{fig_contrastive_network}. Our method can be divided into three steps: 3D model augmentation, line-drawing feature extraction, contrastive loss computation.

\textbf{Step 1: 3D data augmentation.} Our data augmentation happens in 3D instead of 2D \siyuan{because the visual differences between different views are caused by 3D boolean operation}. We generate two sets of images: a branch 1 set and a branch 2 set for each augmented CAD model. Each image set contains the three-view drawings and the isometric view drawing. We denote the branch 1 set by $\left \{F_{1},R_{1},T_{1}, I_{1}\right \}$, and the branch 2 set by $\left\{F_{2},R_{2},T_{2}, I_{2}\right\}$, where $F$, $R$, $T$, and $I$ stand for front, right, top views and the isometric view separately.
The branch 1 and branch 2 image sets are generated from two different modifications to the original CAD model. Each modification is a random Boolean operation (intersection, union, or difference) on the original CAD model with a random primitive (sphere, cube, cone, or cylinder.) 
Figure~\ref{fig_contrastive_network} gives an example of generating the two image sets.

\textbf{Step 2: line-drawing feature extraction.} 
In this paper, $f,g,h$ represent neural networks; $\theta, \phi,\psi$ are the network weights in the three networks respectively. 
$F_i$, $R_i$, $T_i$, and $I_i$ ($i\in \{1,2\}$)
are fed into four convolutional neural networks (CNN) individually. The four networks are denoted by $f_{\theta_1}$, $f_{\theta_2}$, $f_{\theta_3}$, and
$f_{\theta_4}$, where $f_{\theta_j}:\mathbb{R}^{3\times H\times W}\to\mathbb{R}^{C}, j\in \{1,2,3,4\}$; $H$ and $W$ are the height and width of the images. Note that the four networks share the \textit{same architecture} but with \textit{different parameters}. Then, the outputs from $f_{\theta_1}$, $f_{\theta_2}$, and $f_{\theta_3}$ are concatenated and fed into a one-layer MLP $g_{\phi_1}:\mathbb{R}^{3C}\to\mathbb{R}^{K}$, and the output from $f_{\theta_4}$ is fed into another one-layer MLP $g_{\phi_2}:\mathbb{R}^{C}\to\mathbb{R}^{K}$. 
$g_{\phi_1}, g_{\phi_2}$ also share the \textit{same architecture} but with \textit{different parameters}.
The outputs from
$g_{\phi_1}$ and $g_{\phi_2}$ are noted as a $a_i$ and $b_i$ ($i\in \{1,2\}$) , which encode the information from the 3-view images and the isometric image respectively. 

\textbf{Step 3: Contrastive loss computation.} After having the four latent codes ($a_1$, $a_2$, $b_1$, and $b_2$), we concatenate each $a$ and each $b$, which gives four combinations $a_1 \bigoplus b_1$, $a_1 \bigoplus b_2$, $a_2 \bigoplus b_1$, and $a_2 \bigoplus b_2$ (see Figure~\ref{fig_contrastive_network}). 
Then we send the four concatenated latent codes to a binary classifier $h_\psi:\mathbb{R}^{2 K}\to(0, 1)$, where the $h_\psi$ is a two-layer MLP.
The outputs from $h_\psi$ are $\hat{p}_1$, $\hat{p}_2$, $\hat{p}_3$, and $\hat{p}_4$ respectively, and are used to compute the binary cross entropy (BCE) loss with the ground truth. We define the ground truth to be $1$ if the original two latent codes used to concatenate are from the same image pairs, and the ground truth to be $0$ from different image pairs. Therefore, $p_1 = p_4 = 1$, $p_2 = p_3 = 0$.
The final loss is $\frac{1}{4}\sum_{k=1}^{4}BCE(\hat{p}_{k},p_{k})$, ($k\in \{1,2,3,4\}$). 

\textbf{Difference with SimCLR}. Unlike existing contrastive learning methods such as SimCLR, the representation from our method is designed to be both multi-view-consistent and detail-sensitive. Technically, our method differs from them in that: 1) our data augmentation operates on 3D CAD models with Boolean operations (intersection, union, or difference), instead of single-image operations like random cropping, color distortion, and Gaussian blur; 2) our positive pairs are two sets of multi-view line drawings (three-view line drawings and isometric line drawing) that are rendered in the same data branch, and our negative pairs are image sets from different data branches, unlike being sampled from other data instances; 3) we use binary cross-entropy loss to optimize the network, unlike the NT-Xent loss~\cite{gutmann2010noise, zhang2019learning} which has the temperature parameter to tune. \siyuan{Our experimental results also show the advantage of using BCE loss over the NT-Xent loss.}

\siyuan{\textbf{Representation adaption for task \textit{T2I}.}}
We notice that all the learned parameters $\theta, \phi,\psi$ can be loaded to the neural network for further supervised fine-tuning because the contrastive spatial reasoning method is just a pre-training step, and it uses exactly the same network architecture as the network for the supervised learning (see Figure~\ref{fig_contrastive_network} downstream tasks: \textit{T2I}).

\subsection{Self-supervised learning network for task \textit{I2P} and \textit{P2I}}
\begin{figure*}[t]
	\centering
	\includegraphics[width=0.96\linewidth]{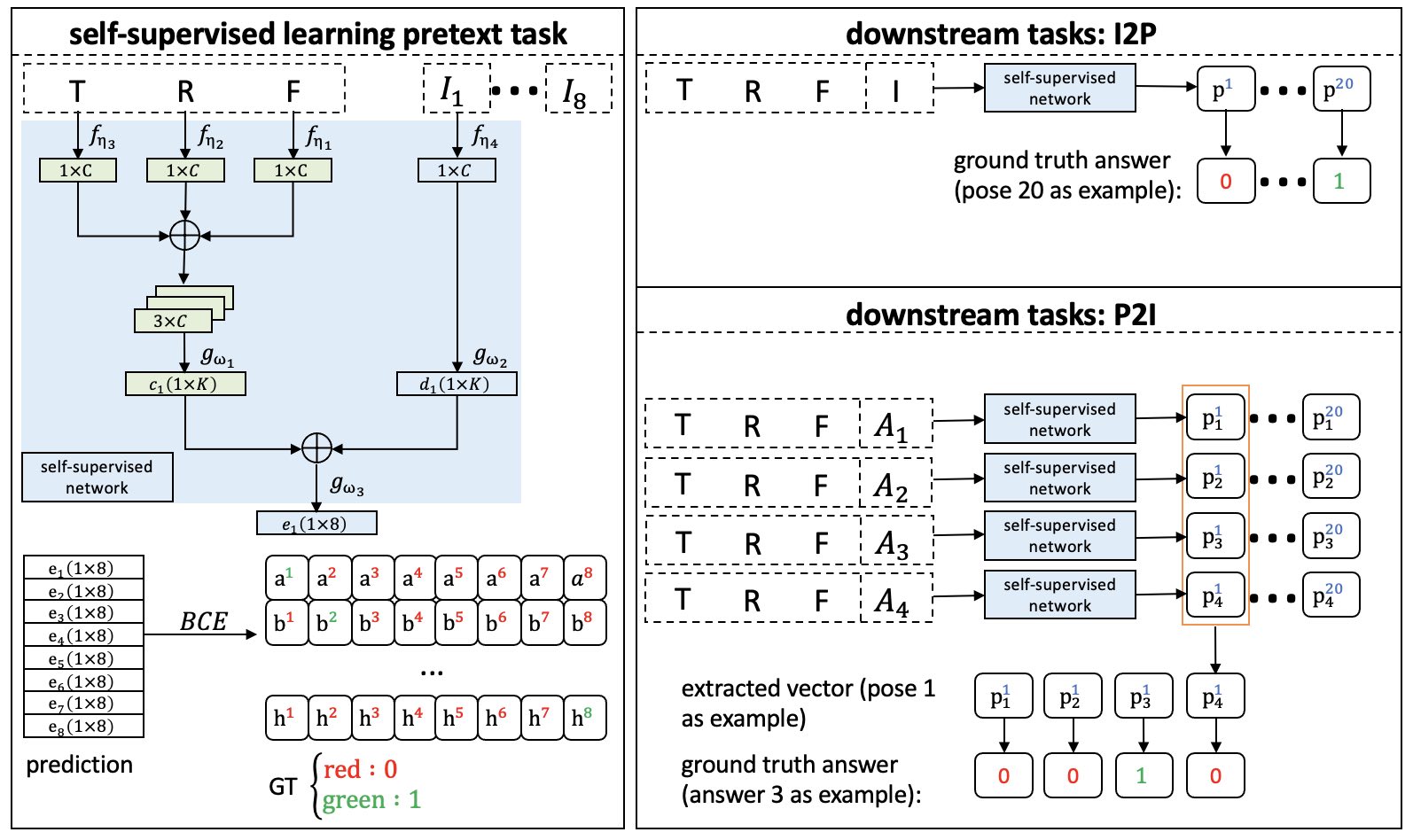}
	\caption{\textbf{Self-supervised learning network architecture for task \textit{I2P} and \textit{P2I}.} We train a self-supervised learning network (left subfigure), then use the learned representations to the downstream tasks \textit{I2P} and \textit{P2I} (right subfigure). The $f_{\eta_1}$, $f_{\eta_2}$, $f_{\eta_3}$, and $f_{\eta_4}$ are CNN networks; $g_{\omega_1}$, $g_{\omega_2}$, and $g_{\omega_3}$ are MLP networks; $c_{1}, d_{1}, e_{1}$ are the encoded feature vectors.}
	\label{fig_self-supervised}
% 	\vspace{-10mm}
\end{figure*}

\siyuan{
We propose a self-supervised learning pretext task, as can be seen in Figure~\ref{fig_self-supervised}. As mentioned in Introduction~\ref{sec:intro}, the learned representations from the pretext task can be helpful to both \textit{I2P} and \textit{P2I} tasks, even when the camera poses in the downstream tasks are not seen in the pretext task. The network design can be divided into two steps: line-drawing feature extraction; loss computation.}

\textbf{Step 1: line-drawing feature extraction.}
\siyuan{
For the three view images, similar as the network design for contrastive spatial reasoning for task \textit{T2I}, $\left \{F,R,T\right \}$ are sent to the share-weight neural network (CNN) $f_{\eta_1}$, $f_{\eta_2}$, $f_{\eta_3}$ separately to obtain three features, where $f_{\eta_j}:\mathbb{R}^{3\times H\times W}\to\mathbb{R}^{C}, j\in \{1,2,3\}$.
Then the three vectors are concatenated and fed into a one-layer MLP $g_{\omega_1}:\mathbb{R}^{3C}\to\mathbb{R}^{K}$, generating $c_{1}$ which encodes the information from the 3-view images.}

\siyuan{
For the eight isometric view images, each of the images goes through the same network. For the sake of simplicity, we use $I_{1}$ (isometric view image rendered from camera pose $1$) as an example. $I_{1}$ is sent to a neural network (CNN) $f_{\eta_4}$ followed by a one-layer MLP $g_{\omega_2}:\mathbb{R}^{C}\to\mathbb{R}^{K}$, and the output latent vector is noted as $d_{1}$.}

\siyuan{
After having $c_{1}$ and $d_{1}$ for three view feature and isometric view feature separately, we concatenate them and send the concatenated vector to a two-layer MLP $g_{\omega_3}:\mathbb{R}^{K}\to\mathbb{R}^{8}$. The obtained 8-dimensional vector $e_{1}$ can represent the camera-pose probability logits of the eight candidate isometric views.  
Therefore, for all the eight isometric view images  $I_{1}, I_{2}, \dots, I_{8}$, we will have eight $1\times8$ vectors, $e_{1}, e_{2}, \dots, e_{8}$ separately. We concatenate the eight vectors to obtain a $8\times8$ matrix, forming the output of the self-supervised learning pretext task.}

\textbf{Step 2: loss computation.}
\siyuan{
This matrix is used to calculate the BCE loss with the ground truth matrix. The ground truth matrix is a $8\times8$ identity matrix, and each row is the logits for the corresponding isometric view. For example, the logit value of the first value in the first row is $1$, while other values are $0$,
since the logits in the first row represents the isometric view pose $1$. The BCE loss is:} \anbang{$\frac{1}{64}\sum_{i=1}^8\sum_{j=1}^8BCE(\hat{p}_{ij},p_{ij}),(i,j\in{1,2,\cdots,8})$, the ground truth $p_{ij}=1$ when $i=j$, otherwise $p_{ij}=0$.}
% Therefore, $f_{\theta_5}$, $f_{\theta_6}$, $f_{\theta_7}$, $f_{\theta_8}$ and $g_{\phi_3}$, $g_{\phi_4}$, $g_{\phi_5}$ forms the self-supervised network, and the learned parameters can be used for downstream task \textit{I2P} and \textit{P2I}.

\textbf{Representation adaption for task \textit{I2P} and \textit{P2I}.}
In task \textit{I2P}, three view images and one isometric view image (rendered from one of the poses in view $1, 2, 5, 6$, pose defined in SPARE3D dataset) are given, and the network is asked to select the correct viewpoint from view $1, 2, 5, 6$. Therefore, for each question, we send the three-view image $F$, $R$, $T$, and the isometric image $I$ to the trained self-supervised network and obtain a eight-digit vector. Then we select the first, second, fifth, and sixth values in the vector as the prediction pose of the network, which can represent the potential view that is used in this task. Then, the max value in the prediction pose vector is considered as the predicted pose, and we give value $1$ to the corresponding pose and $0$ to other poses to form a one hot encoder. Finally, the BCE loss is computed using the predicted one hot encoder and the ground truth one hot encoder.

In task \textit{P2I}, the three-view images and a selected pose from $8$ potential poses are sent to the network. The network then selects the isometric view image corresponding to the pose from four potential answers (four rendered isometric view images). For representation adaption, we send the three-view image $F$, $R$, $T$ and one potential answer image $A_{i}$, $i\in \{1,2,3,4\}$ to the trained self-supervised network and obtain four 8 dimensional vectors. For each vector, the value represents the 
probability of the pose that the answer image is rendered from. Finally, we extract the value from the column that corresponds to the given pose for each vector, forming a four-dimensional vector. For example, the four vectors in the first column will be extracted if the given pose is $1$. Then we use the extracted vector to compute the BCE loss with the ground truth one hot encoder.

\textbf{Extension experiment for camera pose reasoning task.}
\siyuan{Besides testing the self-supervised network's learned representations on \textit{I2P} and \textit{P2I} tasks in the SPARE3D dataset, we are also curious on whether this self-supervised learning could help camera pose reasoning that contains poses unseen during training? Therefore, we design two extension downstream task of \textit{I2P} and \textit{P2I} by increasing the potential camera pose number from $4$, $8$ to $20$, $20$ respectively. 
% The new $12$ camera poses are located at the middle of twelve lines of the cube and pointing to the center. 
The $20$ camera poses are viewed in the supplementary.}

\begin{table*}[t]
\begin{center}
\caption{\textbf{Comparison of performance on \textit{T2I} for SL method vs. SSL method.}
SL and SSL represent supervised learning and self-supervised learning, respectively. 5K and 14K are the training data amount. Fine-tuning means we further use the 5K training data in SPARE3D to fine-tune the parameters.
For supervised learning, we evaluate the network performance for: (1) using \textit{early fusion} or \textit{late fusion} structure (details in the supplementary), (2) whether or not using ImageNet pre-trained parameters.}
% \begingroup
\setlength{\tabcolsep}{1.6pt} % Default value: 6pt
\renewcommand{\arraystretch}{1.5} % Default value: 1
\resizebox{\textwidth}{!}
{%
\begin{tabular}{ c|c|c|c|c|c|c }
\toprule
SL&early-fusion(5K)&early-fusion(pretrained, 5K)&late-fusion(pretrained , 5K)&early-fusion(14K)&early-fusion(pretrained, 14K)&late-fusion(pretrained, 14K)\\
\midrule
&$55.0$&$30.6$&$25.2$&$63.6$&$51.4$&$27.4$\\
\bottomrule
SSL&Jigsaw puzzle\cite{noroozi2016unsupervised}& Colorization\cite{zhang2016colorful} & SimCLR\cite{chen2020simple}& RotNet\cite{gidaris2018unsupervised} & Ours (NT-Xent loss)& Ours (BCE loss)\\
\midrule
& 27.4 & 23.4  & 31.0&30.6 & 48.4 &\textbf{74.9}\\
\bottomrule
\end{tabular}
}

% \endgroup

% For SimCLR and our methods, the supervised learning network cannot be designed as early fusion since it requires two branches for loss computation.}

\label{tab_methods_result}
\end{center}
\vspace{-8mm}
\end{table*}

% \begin{table*}[t]
% \begin{center}
% \caption{\textbf{Comparison of performance on \textit{T2I} for SL method vs. SSL method.}
% SL and SSL represent supervised learning and self-supervised learning, respectively. 5K and 14K are the training data amount. Fine-tuning means we further use the 5K training data in SPARE3D to fine-tune the parameters.
% For supervised learning, we evaluate the network performance for: (1) using \textit{early fusion} or \textit{late fusion} structure (details in the supplementary), (2) whether or not using ImageNet pre-trained parameters.}
% % \begingroup
% \setlength{\tabcolsep}{1.6pt} % Default value: 6pt
% \renewcommand{\arraystretch}{1.5} % Default value: 1
% \resizebox{\textwidth}{!}
% {%
% \begin{tabular}{ c|c|c|c|c|c|c }
% \toprule
% \multirow{2}{*}{SL}
% &early-fusion(5K)&early-fusion(pretrained, 5K)&late-fusion(pretrained , 5K)&early-fusion(14K)&early-fusion(pretrained, 14K)&late-fusion(pretrained, 14K)\midrule\\

% $55.0$&$30.6$&$25.2$&$63.6$&$51.4$&$27.4$\\
% \bottomrule
% SSL&Jigsaw puzzle\cite{noroozi2016unsupervised}& Colorization\cite{zhang2016colorful} & SimCLR\cite{chen2020simple}& RotNet\cite{gidaris2018unsupervised} & Ours (NT-Xent loss)& Ours (BCE loss)\\
% \midrule
% & 27.4 & 23.4  & 31.0&30.6 & 48.4 &\textbf{74.9}\\
% \bottomrule
% \end{tabular}
% }

% % \endgroup

% % For SimCLR and our methods, the supervised learning network cannot be designed as early fusion since it requires two branches for loss computation.}

% \label{tab_methods_result}
% \end{center}
% \vspace{-8mm}
% \end{table*}
\section{Experiments and Discussions}

In this section, we introduce the experiment details for training two of our self-supervised learning networks and the baseline methods.
Then, we compare the classification accuracy of our method with baseline methods on all three tasks, \siyuan{including the extension \textit{I2P} and \textit{P2I} tasks}. 
Additionally, for \textit{T2I} task, we also visualize the attention maps obtained from our methods and supervised baseline methods, demonstrating that our method can better localize the differences between the candidate answers. All experiments are implemented with PyTorch~\cite{PyTorch:NIPS19}, using NVIDIA GeForce GTX 1080 Ti GPU.

\textbf{Data for training and testing.} 
\siyuan{For all the tasks, we use the same test data in SPARE3D dataset for testing (except the \textit{extension I2P} and \textit{extension P2I} tasks). To prevent the network from ``memorize'' the data, we avoid using the same models for self-supervised training, supervised fine-tuning, and testing for all the tasks. Next, we will discuss how we generate training data for different tasks.}

In our contrastive learning approach for \textit{T2I} task, we first generate image sets for two branches. In practice, we download $14,051$ models from the ABC dataset for training and $737$ models for testing. Then we use PythonOCC~\cite{pythonocc} to apply different random Boolean operations on each model, generating four images for branch 1 and branch 2, respectively. Therefore, we have $14,051 \times 8$ rendered images. 

All $14,051 \times 8$ images are used for three self-supervised baselines that we compare with, namely Jigsaw Puzzle, Colorization, SimCLR, and RotNet.
For supervised learning with 5K dataset, we use the original data from SPARE3D paper, with $5,000$ questions for training. For supervised learning with 14K dataset, we use the $14,051$ CAD models to generate new questions for training.

For the self-supervised network for \textit{I2P} and \textit{P2I}, we generate different training sets with CAD model amount ranging from $5,000$ to $40,000$, with $5,000$ as a step. For each CAD model, it generates three-view images($F, R, T$) and eight isometric view images($I_1,I_2,\cdots,I_8$). For a fair comparison, we also generate different scale datasets for supervised learning, ranging from $5,000$ to $40,000$, and all the implementation settings are the same as in SPARE3D. 

For both the \textit{extension I2P} and \textit{extension P2I} tasks, we need to: 1) create supervised training dataset with different size, ranging from $5,000$ to $40,000$, and 2) create a test set with $1,000$ questions, which is the same number as in the original \textit{I2P} and \textit{P2I} settings. We follow the implementation settings as in SPARE3D, and the only change is we increase the potential view number to $20$.

\textbf{Hyperparameter settings.}
We tune the learning rate and batch size for each self-supervised learning method and supervised learning method for all tasks. 

For \textit{T2I}, our contrastive learning network, with either NT-Xent loss or BCE loss, uses a learning rate of $0.00005$, batch size $4$. Jigsaw puzzle, Colorization, SimCLR, and RotNet all use learning rate $0.00001$ and batch size $10, 30, 70, 16$, respectively. For supervised learning, we follow the hyperparameter settings as in SPARE3D.

For the \textit{I2P} and \textit{P2I} tasks, the learning rate and batch size for our self-supervised network are $0.00005$ and $4$. \anbang{ For the \textit{extension I2P} and \textit{extension P2I} tasks, the learning rate is $0.00001$, and the batch size are $70$ and $20$ respectively.} For supervised learning, we follow the hyperparameter settings as in SPARE3D. 

\textbf{Details of the two proposed self-supervised networks.} For both our contrastive learning method and self-supervised method, we use the VGG-16 network as the backbone for image feature extraction, and parameters keep the same. Note that for the contrastive learning network using NT-Xent loss, we use $a_{1}, b_{1}, a_{2}, b_{2}$ as the latent vectors. $a_{1}, b_{1}$ and $a_{2}, b_{2}$ are considered as positive pairs, while other remaining pairs, including the pairs within the batch, are negative. The value of $C$ is set to be $18,432$ (by flattening the $6 \times 6 \times 512$ feature map).
The value of $K$ is set to be $4096$, the same as in the SPARE3D paper. The size of input line drawings is $200 \times 200 \times 3$. The output of the last convolutional layer is a $6 \times 6 \times 512$ feature map.

\textbf{Self-supervised learning baseline network adaptation for \textit{T2I}.}
For task \textit{T2I}, we use Jigsaw puzzle~\cite{noroozi2016unsupervised}, Colorization~\cite{zhang2016colorful}, SimCLR~\cite{chen2020simple}, and RotNet~\cite{gidaris2018unsupervised} as four self-supervised learning baseline methods to compare with our method, see Table~\ref{tab_methods_result}. For the network structures of the Jigsaw puzzle, Colorization, and RotNet, we follow these papers' original design, only replacing the backbone networks with VGG-16 to ensure that the learned parameters can be loaded to the networks of the downstream tasks. 
For SimCLR, we define the positive ``image pairs'' in our case as the $\left \{F,R,T\right \}$ and $I$ images, which are generated from the same CAD model. We then use the same \textit{contrastive network} as in Figure~\ref{fig_contrastive_network} to extract the features for $\left \{F,R,T\right \}$ and $I$ images separately, and the learned parameters can be loaded to the downstream tasks.

% For the Jigsaw puzzle and Colorization, the VGG-16 backbone network takes one single image as the input. Therefore, on the one hand, we can load the learned parameters $\theta$ for supervised fine-tuning with the \textit{early fusion}, discarding the first convolutional layer's parameters. On the other hand, we can load the learned parameters $\theta$ for supervised fine-tuning with the \textit{late fusion} by initializing the parameters of the four $f$ networks with the same pre-trained parameters. 
\textbf{\textit{Extension I2P} and \textit{extension P2I} task implementation.} For these two tasks that use $20$ views as potential camera pose, we compare our self-supervised learning pre-trained networks with the supervised learning baseline methods. For supervised learning baseline methods of \textit{extension I2P} and \textit{extension P2I}, we follow the network design as \textit{I2P} and \textit{P2I} respectively. The only difference is that, for each task, the output of the last MLP layer $g_{\omega_3}$ changes to a $20$ dimensional latent vector. Therefore, when loading the learned parameters from the self-supervised network, we discard the last layer MLP of the trained model and train the parameters using the extension supervised learning training data for 20 views.
 
% For using the learned representations from the self-supervised network, we load the parameters from the trained model for all the layers in the downstream network, and the downstream network uses the same structure as supervised learning.

% \anbang{For I2P and P2I downstream tasks for 20 views, we only change the last layer of MLP from $1\times8$ to $1\times20$. In SR-pretrained experiment, we load trained self-supervised network's parameters for all other layers, while for ImageNet-pretrained experiment, we load Image-Net pretrained parameters for all layers except first and last layers of VGG16.}

\begin{figure*}[t]
	\centering
	\includegraphics[width=1\textwidth]{figs/Picture5.jpg}
	\caption{\textbf{Attention maps for SL vs. SSL method in \textit{T2I} task.} 
	For each CAD model, the first row are the line drawings. The second and third row are the attention maps generated from supervised learning using \textit{early fusion} and \textit{late fusion}, respectively. The fourth row are the attention maps generated from our method. N$/$A indicates no attention map for the corresponding view. Best viewed in color.
	}
	\label{fig_attn_maps}
	\vspace{-5mm}
\end{figure*}
\vspace{-1mm}

\subsection{\siyuan{\textit{T2I} task result analysis}}

\textbf{Classification accuracy for task \textit{T2I}.}
As can be seen in Table~\ref{tab_methods_result}, our methods (both with or without fine-tuning) outperform other methods, including self-supervised baseline methods and supervised methods. Our fine-tuned result can achieve $74.9\%$ accuracy on \textit{T2I} task, approaching the average untrained human performance of $80.5\%$.
Here direct evaluation means we use the learned parameters from the trained contrastive learning network, and fine-tuning means we further use the $5,000$ training data for supervised learning to fine-tune the learned parameters.

Although we use more data in the contrastive pre-training,
the higher accuracy of our method is not only due to increased data volume. As aforementioned, we use $14,051$ CAD models to generate image sets for contrastive learning. We also use these models to generate $14,051$ questions for purely supervised learning. This ensures the number of CAD models used for our method is the same as for purely supervised learning.
With the same number of CAD models for training (14K dataset), we find the best performance that supervised learning can achieve is $63.6\%$. Although increasing the data volume can help improve the baseline performance for supervised learning, from $55.0\%$ to $63.6\%$, the result is still significantly lower than our method, which is $74.9\%$. 

We believe the good performance of our method is because contrastive learning helps the network learn the \emph{detail-sensitive} yet view-invariant visual representations in the line drawings. 
This reason could also explain an interesting phenomenon that we observe, which is that the direct evaluation (without fine-tuning) using the learned parameters from contrastive spatial reasoning can achieve $71.4\%$ accuracy (see Table~\ref{tab_methods_result}). Thus, a good visual representation should be able to transfer to the downstream tasks with little further training.

Qualitatively, we show that our contrastive learning method can help the network learn the \emph{detail-sensitive} yet view-invariant visual representations.
we visualize the attention map for our contrastive learning method and supervised learning method using the schemes in \cite{zagoruyko2016paying}. {For supervised learning, we generate attention maps on both \textit{early fusion} and \textit{late fusion} method with no pre-trained parameters. For \textit{early fusion}, the input line drawings, including front, right, top, and one isometric line drawing, are concatenated before being sent to the CNN. Therefore, each composite image will have one attention map. We put the attention map with the corresponding candidate isometric line drawing, leaving the attention map for the front, right, and top as empty. For \textit{late fusion} and our method, after having the attention map for each input image, we put it together with the corresponding input line drawing. All the results are shown in Figure~\ref{fig_attn_maps} (more results in the supplementary).} The comparison between the three rows of attention maps generated from the three methods shows that our method can help the CNN better capture the tiny detailed differences between the candidate answer drawings, which is the key to selecting the correct answer from four similar options.
\begin{table}[t]
\begin{center}
\caption{\textbf{Comparison of performance on \textit{I2P} and \textit{P2I} tasks for SL method vs. SSL method.} }
\vspace{-3mm}
% \begingroup
\resizebox{\linewidth}{!}{%
\begin{tabular}{ c|c|c|c|c|c|c|c|c } 
\toprule
 Data amount (K)&5&10&15&20&25&30&35&40\\
\midrule
I2P(SL)& 83.6&	86.4&	87.7&	88.5&	88.7&	90.4&	90.6&	91.1\\

I2P(SSL)& 88.7&	93.2&	95.1&	96.4&	96.7&	97.7&	97.5&
\textbf{98.0}\\

P2I(SL)&65.4&67.1&68.5&67.8&69.8&69.6&68.5&70.4\\%\multicolumn{8}{c|}{All around 25}\\

P2I(SSL)&72.4&	80.8&	81.9&	82.1&	82.8&	83.1&	83&	\textbf{83.4}\\
\bottomrule
\end{tabular}
}

% \endgroup

\label{tab_I2P_P2I}
\end{center}
% \vspace{-4mm}
\end{table}

\begin{table}[t]
\begin{center}
\caption{\textbf{Comparison of performance on \textit{extension I2P} and \textit{extension P2I} tasks for SL method vs. SSL method.} Note that for SSL methods, we fine-tune the network for the downstream tasks.
}
\vspace{-3mm}
% \begingroup
\resizebox{\linewidth}{!}{%
\begin{tabular}{ c|c|c|c|c|c|c|c|c } 
\toprule
         Data amount(K)&5&10&15&20&25&30&35&40  \\
    \midrule
         I2P(SL)&15.1&14.5&15.9&14.8&17.7&17.8&16.3&17.5\\
         I2P(SSL)&42.9&45.5&48.4&50.1&51.3&51.7&54.8&\textbf{56.5}\\

        P2I(SL)&51.4&47.2&33.7&52.9&44.3&43.3&52.7&44.8\\

        P2I(SSL)&67.4&73.5&66.2&76.2&75.0&68.1&75.7&\textbf{77.9}\\
    \bottomrule
\end{tabular}
}

\label{I2P_SR_downstream}
\end{center}
\vspace{-3mm}
\end{table}

% \textit{Imagenet pre-training helps contrastive spatial reasoning network converge.} We find it a necessity to use the Imagenet pre-trained parameters to initialize the VGG-16 network for our method. Otherwise, it cannot converge. Without the pre-trained parameters, the training and testing accuracy of contrastive learning is around $25\%$, showing the network solves the problem using random guesses. It coincides with the conclusion from the controlled experiment(see in supplementary) that Imagenet pre-training has an influence on the network's performance. It seems that although these learned parameters are pre-trained on the Imagenet dataset for the image classification problem, which is a different domain of solving spatial reasoning tasks using line drawings, they can still help contrastive learning.
\subsection{\siyuan{\textit{I2P} and \textit{P2I} task result analysis}}

\textbf{Classification accuracy for task \textit{I2P} and \textit{P2I}.} 
For a fair comparison, we change the supervised learning network's structure to match the structure of our self-supervised learning method. More details will be discussed in the supplementary.
The results can be seen in Table~\ref{tab_I2P_P2I}. With the increase of data amount for training, both supervised learning, and our self-supervised learning-based method achieve higher accuracy. For task \textit{I2P}, the best accuracy achieves $98.0 \%$, and for task \textit{P2I}, the best accuracy is $83.4 \%$. The best performance happens when using the $40,000$ scale dataset with our self-supervised learning method. 

Therefore, we claim that the increased data amount can help both supervised and self-supervised learning methods. However, our self-supervised learning method has the advantages that: 1) more efficient to improve the accuracy, compared with using the same amount of data as supervised learning methods, and 2) more helpful when the supervised learning method does not perform well. For the second advantage, since the number of potential cameras poses in task \textit{I2P} and \textit{P2I} are $4$ and $8$ respectively, it is natural to expect that it will be more difficult for the neural network to solve the \textit{P2I} task. 
In \textit{P2I} task, we find that even after increasing the training data amount for supervised learning, the performance improvement is limited.
However, our self-supervised learning method can achieve around $10 \%$ higher accuracy than the supervised learning method, revealing its advantage over the supervised learning method.

\textbf{Classification accuracy for task \textit{extension I2P} and \textit{extension P2I}.} 
We also show that our self-supervised learning methods can help \textit{extension I2P} and \textit{extension P2I}. Table~\ref{I2P_SR_downstream} shows our self-supervised learning-based method outperforms the supervised learning method for different data amounts on both tasks. We believe it is because the ``rough'' pose reasoning in the pretext task can help the ``fine'' pose reasoning in the downstream task. ``rough'' is because the network only needs to reason about $8$ poses; ``fine'' means in the downstream tasks, the network is required to reason about more views, and these views are located in 
between the $8$ poses. Once the network has the ability to determine the ``rough'' pose and use it as prior information, it will be easier for the network to reason about the ``fine'' pose.

% We also explore the best VGG network as the backbone for our method. As can be seen from Table~\ref{tab_ablation_backbone}, we find VGG-19 can provide the best performance among three different types of VGG networks.

% \begin{table}[t]
% \begin{center}
% \begin{tabular}{ c|c|c|c } 
% \hline
% backbone & VGG-13 & VGG-16 & VGG-19 \\
% \hline
% acc & 72.0 & 74.9 & \textbf{75.0} \\
% \hline
% \end{tabular}
% \caption{\textbf{Comparison of using different VGG backbones}. The results show that VGG-19 is most suitable for our method.}
% \label{tab_ablation_backbone}
% \vspace{-6mm}
% \end{center}
% \end{table}

% \vspace{-3mm}

\subsection{Limitations and discussion}
\siyuan{One limitation is that our methods are designed for the tasks in the SPARE3D dataset. Therefore, they are beneficial for \textit{object level} spatial reasoning, yet not directly for \textit{scene level} spatial reasoning. One main difference between these two settings is whether the views are ``outside-in'' for 3D shapes or ``inside-out'' for scenes~\cite{esteves2019equivariant}. To make spatial reasoning helpful in robot manipulation, navigation or other scenarios,  \textit{scene level} spatial reasoning is necessary.}

\siyuan{Besides, our methods are tested on a specific dataset and solve some basic spatial reasoning tasks. More effort is required to explore how our methods can be applied to more general spatial reasoning tasks, which can help the community better utilize the power of spatial reasoning to solve real-world problems.}  

% \textbf{Potential negative social impacts.}
% Although we believe our methods will not lead to negative social impacts with respect to the research domain, we still admit that developing deep learning methods will consume much power, which has a negative impact on climate change.
% \vspace{-1mm}
\section{Conclusion}\label{sec:conclusion}\vspace{-2mm}
In this paper, we focus on enhancing the deep networks' performance on multi-view line drawing spatial reasoning tasks on the SPARE3D dataset. Specifically, we focus on two types of tasks: 1) view consistency task, which contains task \textit{T2I}, 2) camera pose reasoning task, which contains task \textit{I2P} and \textit{P2I}. We quantitatively and qualitatively show the advantage of using our self-supervised learning methods for all three tasks.

% For task \textit{T2I}, we propose a simple yet effective contrastive learning method for line drawings in SPARE3D.
% More specifically, our method outperforms the prior method in SPARE3D by a large margin, and it shows more reasonable attention maps. We conjecture that our method is effective because it can effectively learn detail-sensitive yet view-invariant representations.
% For task \textit{I2P} and \textit{P2I}, we propose a self-supervised learning method that can help the network find correlations between the pose and the image that is rendered from that pose. Our experiment results show that our self-supervised learning method always outperforms the previous supervised learning method when using the same amount of data for training. Moreover, we find our pre-trained pretext task model can help the downstream tasks even their camera poses are not seen in the pretext task, revealing the effectiveness of our pre-trained model. 

\siyuan{In the future, We plan to explore how our network design for reasoning about 2D and 3D information could benefit the more general vision-related tasks like AI-assisted design, localization, and navigation. }

% Overall, we believe it is promising to use self-supervised learning methods for multi-view drawings to solve spatial reasoning tasks.
% \vspace{-5mm}

% \small{
% \section*{Acknowledgment}
% The research is supported by the NSF Future Manufacturing program under EEC-2036870. Siyuan Xiang gratefully thanks the IDC Foundation for its scholarship. We also thank the anonymous reviewers for their constructive feedback.
% }

\section*{Acknowledgment}
The research is supported by NSF FM Program under EEC-2036870. Siyuan Xiang gratefully thanks the IDC Foundation for its scholarship. The authors gratefully acknowledge the constructive comments and suggestions from the anonymous reviewers and area chairs.
{\small
\bibliographystyle{ieee_fullname}
\bibliography{egbib}
}
\end{document}

\def\cvprPaperID{11115}

% %%%%%%%%% TITLE
\title{Self-supervised Spatial Reasoning on Multi-View Line Drawings}

\author{
 {Siyuan Xiang\thanks{Equal contribution.} \ \footnotemark[3]{}}
 \and
 {Anbang Yang\footnotemark[1]{} \ \footnotemark[3]{}}
 \and
 {Yanfei Xue\footnotemark[3]{}}
 \and
 {Yaoqing Yang\footnotemark[4]{}}
 \and
 {Chen Feng\thanks{The corresponding author is Chen Feng {\tt\small cfeng@nyu.edu}.} \ \footnotemark[3]{}}
 \and
 \normalsize{\textsuperscript{$\ddagger$}New York University Tandon School of Engineering \quad \textsuperscript{$\mathsection$}University of California, Berkeley}
 \\ \url{https://ai4ce.github.io/Self-Supervised-SPARE3D/}
}

\maketitle
\appendix

\section*{Appendix}

\section{Supervised Learning Exploration}

We have explored three factors that might affect the performance of supervised learning based methods: (1) the quality of data used for training, (2) the network's capacity (width and depth), and (3) the network's structure. We show the influence of factor 1 in the paper, and we will show the influences of factor 2 and 3 in the supplementary. Since task \textit{I2P} and \textit{P2I} both belong to \textit{camera pose reasoning} task, we only test the factor influence on \textit{I2P} for the sake of limited computational resources.

\subsection{Network's Capacity Exploration}

\textit{Network depth for \textit{T2I} and \textit{I2P} tasks.} The depth of the network represents the number of layers of the VGG family backbone networks. We use VGG-13, VGG-19 based backbone to represent three different depths of the network.

\textit{Network width for \textit{T2I} and \textit{I2P} tasks.} Table~\ref{tab_supp_network_width_details} shows the detailed network width for all VGG-16 based backbone network with different widths, for \textit{T2I} and \textit{I2P} tasks. 

\begin{figure}[h  t]
	\centering
	\includegraphics[width=1\columnwidth]{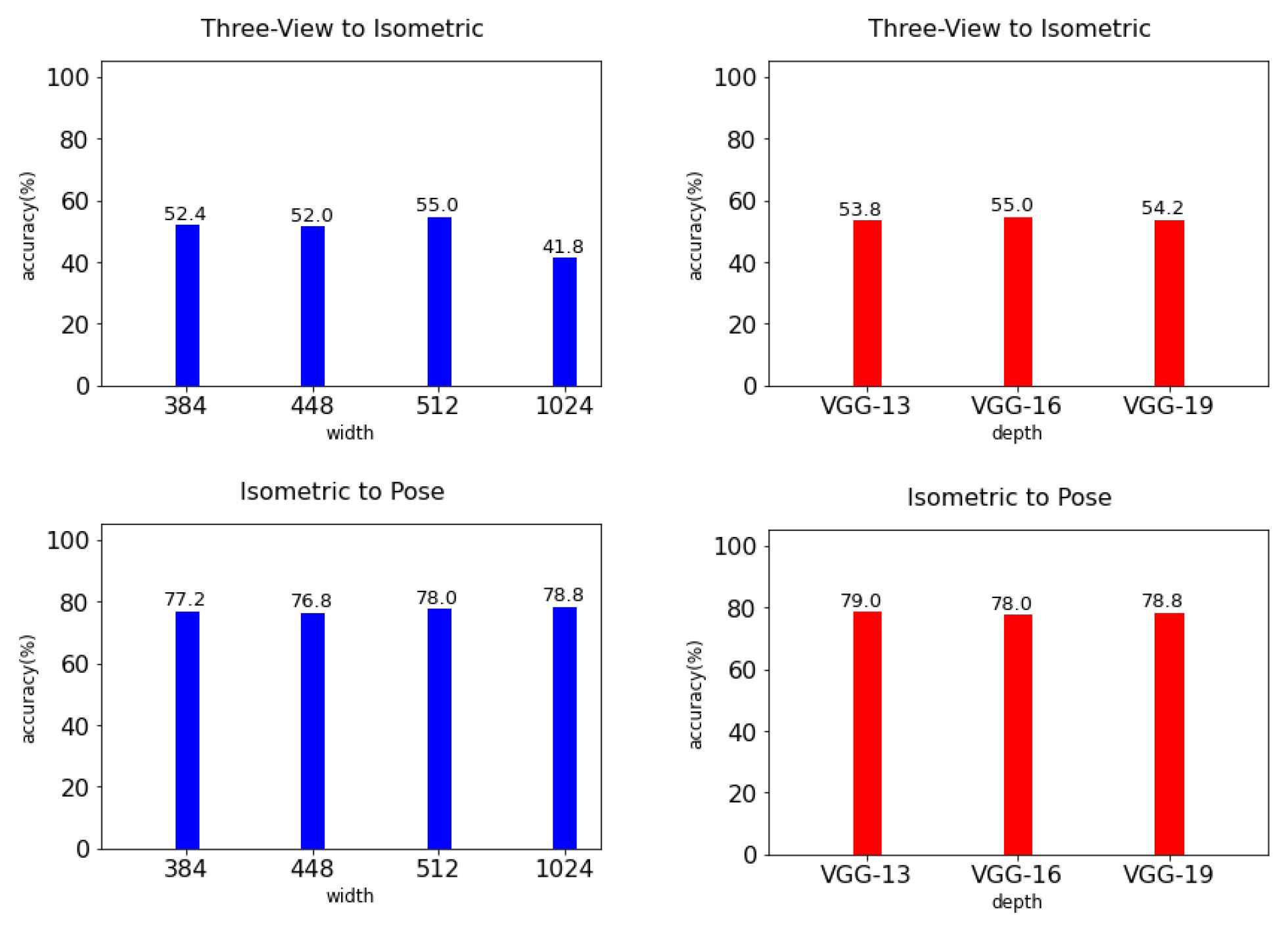}
	\vspace{-2mm}
	\caption{\textbf{Network capacity (width and depth) vs. test accuracy}. The results show that naively increasing the network capacity cannot improve the network performance on \textit{T2I} and \textit{I2P} tasks.}
	\label{fig_network_capacity}
	\vspace{-1mm}
\end{figure}

For task \textit{T2I}, decreasing the network's width does not hurt the network's performance, although strangely, increasing the network's width leads to a decrease in the testing accuracy (Figure~\ref{fig_network_capacity} top-left). For the depth control experiments, we find almost no differences among the three selected network depth values (Figure~\ref{fig_network_capacity} top-right). 
For task \textit{I2P}, neither the width nor the depth of the network can significantly improve the network's performance(Figure~\ref{fig_network_capacity} bottom).

\begin{table}[ht]
\begin{center}
\begin{tabular}{ c|c|c|c|c } 
 \hline 
 $\!$ & $384$ & $448$ & $512$ & $1024$ \\ 
 \hline
 $1-2$ & $48$ & $56$ & $64$ & $128$ \\ 
 $3-4$ & $96$ & $112$ & $128$ & $256$ \\ 
 $5-7$ & $192$ & $224$ & $256$ & $512$ \\ 
 $8-13$ & $384$ & $448$ & $512$ & $1024$ \\ 
 \hline
\end{tabular}
\caption{\textbf{Network width for VGG-16 based backbone network.} $384$, $448$, $512$, $1024$ are the width of the modified VGG-16 backbone network. $1-2$, $3-4$, $5-7$, $8-13$ means convolutional layer $1-2$, $3-4$, $5-7$, $8-13$ respectively. The number in each cell means the width of that convolutional layer (the number of channels).}
\label{tab_supp_network_width_details}
\end{center}
% \vspace{-2mm}
\end{table}

\subsubsection{Network Structure Exploration}

\textit{P2I network structure.} As mentioned in our paper, to make a fair comparison, we modify the supervised baseline method for the \textit{P2I} task so that it has the same network structure for feature extraction as our self-supervised network. In this section, we introduce the details of the modified network structure for the \textit{P2I} task. 

Each question in \textit{P2I} task contains the 3-view(front, right, top view) line drawings and one given pose out of eight views. We note these three drawings as $F$, $R$, $T$, and the given pose as $P_{m}$, ($m\in \{1,2,3,4,5,6,7,8\}$). The answers provide four candidate drawings. These candidate drawings are isometric view line drawings rendered from four views, which are randomly selected from eight views designed in SPARE3D dataset. We note the four candidate answers as $I_{1}$, $I_{2}, I_{3}, I_{4}$. For the \textit{early fusion method}, we concatenate $F$, $R$, $T$ and one of the isometric drawings $I_{i}$, ($i\in \{1,2,3,4\}$) to form a $12-channel$ composite image $I_{c_{i}}$, ($i\in \{1,2,3,4\}$). 
Then, we send $I_{c_{i}}$ to a VGG-based classifier: $g_{\theta}:\mathbb{R}^{12\times\ H\times W}\to\mathbb{R}^{8}$, where $\theta$ represents the parameters in the network. The $8$ number codeword represents the probability of the composed image $I_{c_{i}}$ belonging to eight coded views. Then, we pick the number of the codeword corresponding to view the $P_{m}$, note as $\hat{p}_{m_{i}}$, ($i\in \{1,2,3,4\}$). 
The ground truth is set to be $1$ if the candidate isometric drawing is rendered from view $P_{m}$, and otherwise $0$.
With the provided ground truth $p_{m}$, we can compute the BCE loss to train the neural network, which is: $\frac{1}{4}\sum_{k=1}^{4}BCE(\hat{p}_{m_{k}},p_{m_{k}})$. 

\textit{Network structure for all tasks.}
\begin{table*}[ht]
\begin{center}
\begin{tabular}{ c|c|c|c|c|c|c|c|c } 
\hline
index & pre-train & late fusion & no pooling & no dropout & share weight & separate fc & max & avg\\
\hline
\rowcolor{GreenYellow}
1 & n & n & n & n & n & n & $76.8$ & $72.5$\\ 
2 & y & n & n & n & n & n & $80.4$ & $78.9$\\
\hline

\rowcolor{GreenYellow}
3 & n & n & y & n & n & n & $76.4$ & $71.1$\\
4 & y & n & y & n & n & n & $79.4$ & $79.1$\\
\hline

\rowcolor{GreenYellow}
5 & n & n & n & y & n & n & $77.8$ & $69.1$\\
6 & y & n & n & y & n & n & $81.2$ & $80.9$\\
\hline

\rowcolor{GreenYellow}
7 & n & n & y & y & n & n & $70.4$ & $66.8$\\ 
8 & y & n & y & y & n & n & $80.8$ & $79.1$\\ 
\hline

\rowcolor{GreenYellow}
9 & n & y & y & y & n & n & $78.8$ & $77.4$\\ 
10 & y & y & y & y & n & n & \textbf{86.4} & $84.1$\\ 
\hline

\rowcolor{GreenYellow}
11 & n & y & y & y & n & y & $80.0$ & $76.0$\\ 
12 & y & y & y & y & n & y & $85.4$ & $84.3$\\
\hline

\rowcolor{GreenYellow}
13 & n & y & y & y & y & y & $75.6$ & $74.2$\\ 
14 & y & y & y & y & y & y & $85.4$ & $84.9$\\
\hline

\end{tabular}
\caption{\textbf{Network architecture vs. performance on \textit{I2P} task.} The backbone used is VGG-16. The rows in the green background represent the networks are not initialized with the Imagenet pre-trained parameter, while the rows in the white background are initialized with the parameters. Every two rows (odd row and even row) can be compared to see the influence of using pre-trained parameters or not. We provide the max and average results for each type of network based on seven times of implementation.}
\vspace{5mm}
\label{tab_supp_network_structure}
\end{center}
% \vspace{-5mm}
\end{table*}
As we mentioned in the paper, we explore many variants of the baseline network structure, to see if a variant will affect the network's performance on the task. Because of the limitation of our computational resources, we explore the variants on the task \textit{I2P}.
Here, we list out all the six variants we tried and the results in Table~\ref{tab_supp_network_structure}. Among them, two variants have a consistent impact on the tasks, which are (1) whether using pre-trained parameters from ImageNet, (2) using \textit{early fusion} or \textit{late fusion} for image feature extraction. 

\textit{Early fusion} vs. \textit{late fusion}. In the original SPARE3D paper, the baseline backbone network treats all the input images (front, right, top view drawings, and one isometric view drawing from the candidate answers) as a whole, and it concatenates those images before sending them to the first convolutional layer. We call this way of feeding multi-view line drawings to a network as the \textit{early fusion}. In contrast, we design a network that takes the three-view drawings and the isometric view drawing as separate inputs, which means the input drawings are sent to a convolutional network that shares the \textit{same architecture} 
yet has \textit{separate network parameters}. We name this way 
of separately handling the input as the \textit{late fusion} since the extracted image features are concatenated later. Other network structures are kept the same as the baseline method in SPARE3D.

\textit{Pre-training} vs. \textit{No pre-training}. As in many other research works, we find using ImageNet pre-trained parameters for the backbone VGG network has obvious influnece on our tasks. 

Next, we will focus on the remaining four variants that do not have obvious influence on our tasks: (1)``no pooling'', (2)``no dropout'', (3)``share weight'', and (4)``separate fc'' respectively. ``no pooling'' means we discard all the adaptive average pooling layer in the VGG-16 backbone. ``no dropout'' means we delete all the dropout layers in the VGG-16 backbone. ``share weight'' means for the \textit{late fusion method}, all the VGG-16 backbone use the \textit{same architecture} and with \textit{same parameters}. ``separate fc'' means for the \textit{late fusion method}, the front, right, top view drawings are first fed into the VGG-16 backbone based network: $g_{\phi}:\mathbb{R}^{3\times\ H\times W}\to\mathbb{R}^{18432}$. We note the image features as $c_{f}, c_{r}, c_{t}$ separately. Then we concatenate the three codewords to form a code word, and maps it via an MLP: $g_{\psi}:\mathbb{R}^{55296}\to\mathbb{R}^{18432}$. For the isometric drawing, we send it to the VGG-16 backbone based network: $g_{\phi}:\mathbb{R}^{1\times\ H\times W}\to\mathbb{R}^{18432}$. Finally, we concatenate the two codewords generated from 3-view drawings and isometric drawing as the feature vector for the classification. Other parts of the network are the same as not using the ``separate fc'' structure.

Table~\ref{tab_supp_network_structure} reveals that ``no pooling'', ``no dropout'', ``share weight'', ``separate fc'' has no obvious and consistent impact on the network's performance. Since rows with an odd number as index differ from the rows with even numbers in ``pre-train'', each time we will compare two odd rows, which are not using ``pre-train''. The same conclusion can be drawn if we compare two even rows each time.

We can compare the $1$ row with the $3$ row, and we can find that ``no pooling'' does not obviously affect the results. If we compare the $1$ row with the $5$ row, we can find ``no dropout'' also cannot help the network perform better. Comparing the $1$ row with the $7$ row, we have the conclusion that using both ``no pooling'' and ``no dropout'' could not improve the network's classification accuracy.

For $9,11,13$ rows, we use the \textit{late fusion method}. Based on this method, we vary the network's structure of the remaining four variants. We also find these four variants do not have a significant influence on the network for \textit{late fusion method}. For $9$ row, ``no pooling'' and ``no dropout'' seem not to impact on the classification results; for $11$ row, ``no pooling'', ``no dropout'', and ``separate fc'' does not work; for $13$ row, all the four variants cannot help improve the performance.

Therefore, we conclude that except for the two factors we mentioned in the previous section, the other four factors do not have an obvious impact on the final classification results on the \textit{I2P} task.

\section{Twenty Camera Poses for Extension Tasks}
\begin{figure}[t]
	\centering
	\includegraphics[width=0.45\textwidth]{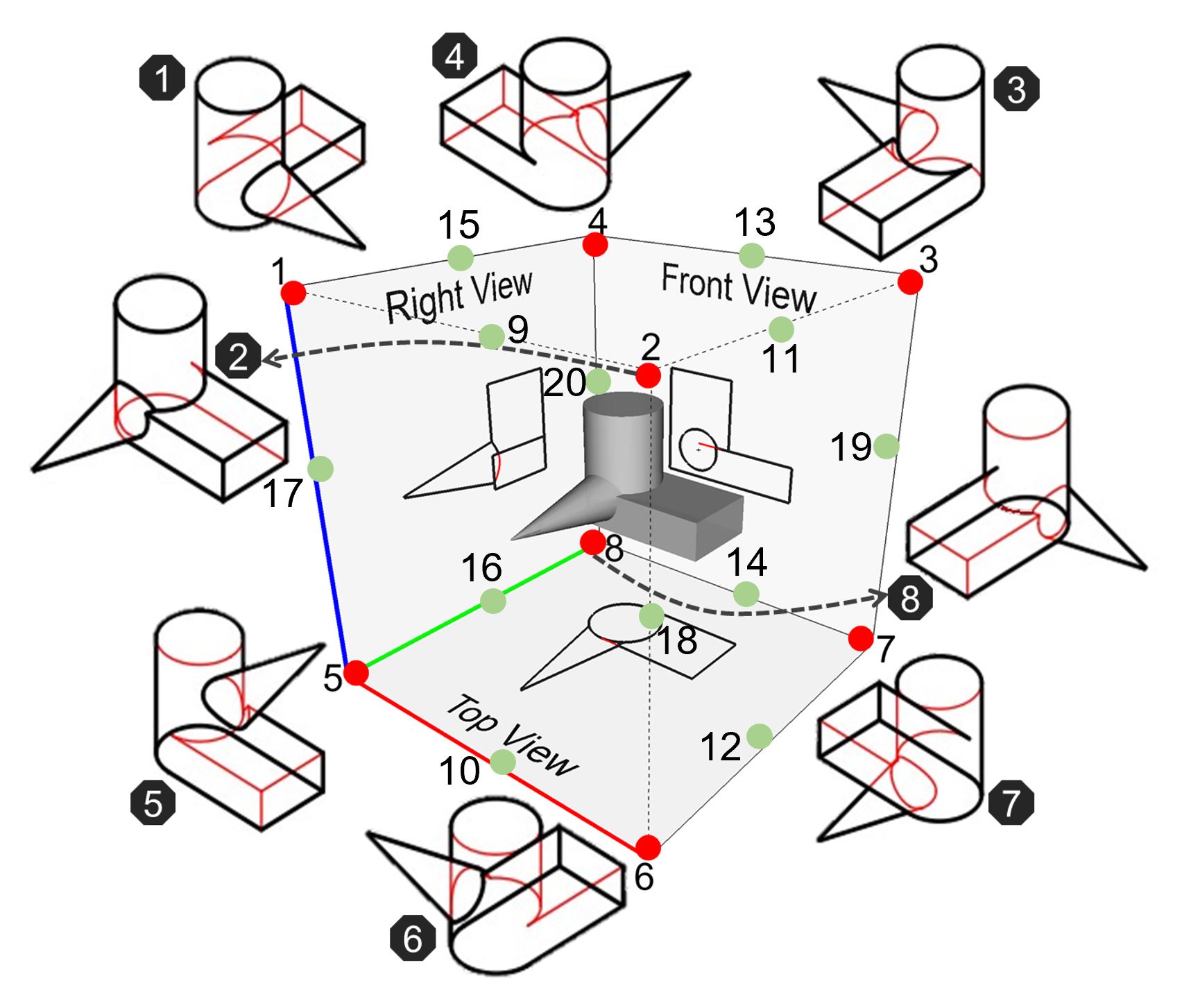}
\caption{\textbf{Twenty camera poses for \textit{extension I2P} and \textit{extension P2I} tasks.}}
\label{fig_20_poses}
\end{figure}
As we mentioned in our paper, we extend the \textit{I2P} and \textit{P2I} tasks to \textit{extension I2P} and \textit{extension P2I} using twelve more camera poses. We show all the twenty camera poses in Figure~\ref{fig_20_poses}.

\begin{figure*}[t]
	\centering
	\includegraphics[width=1\textwidth, trim=.0cm 18cm 5.0cm 0.0cm,clip]{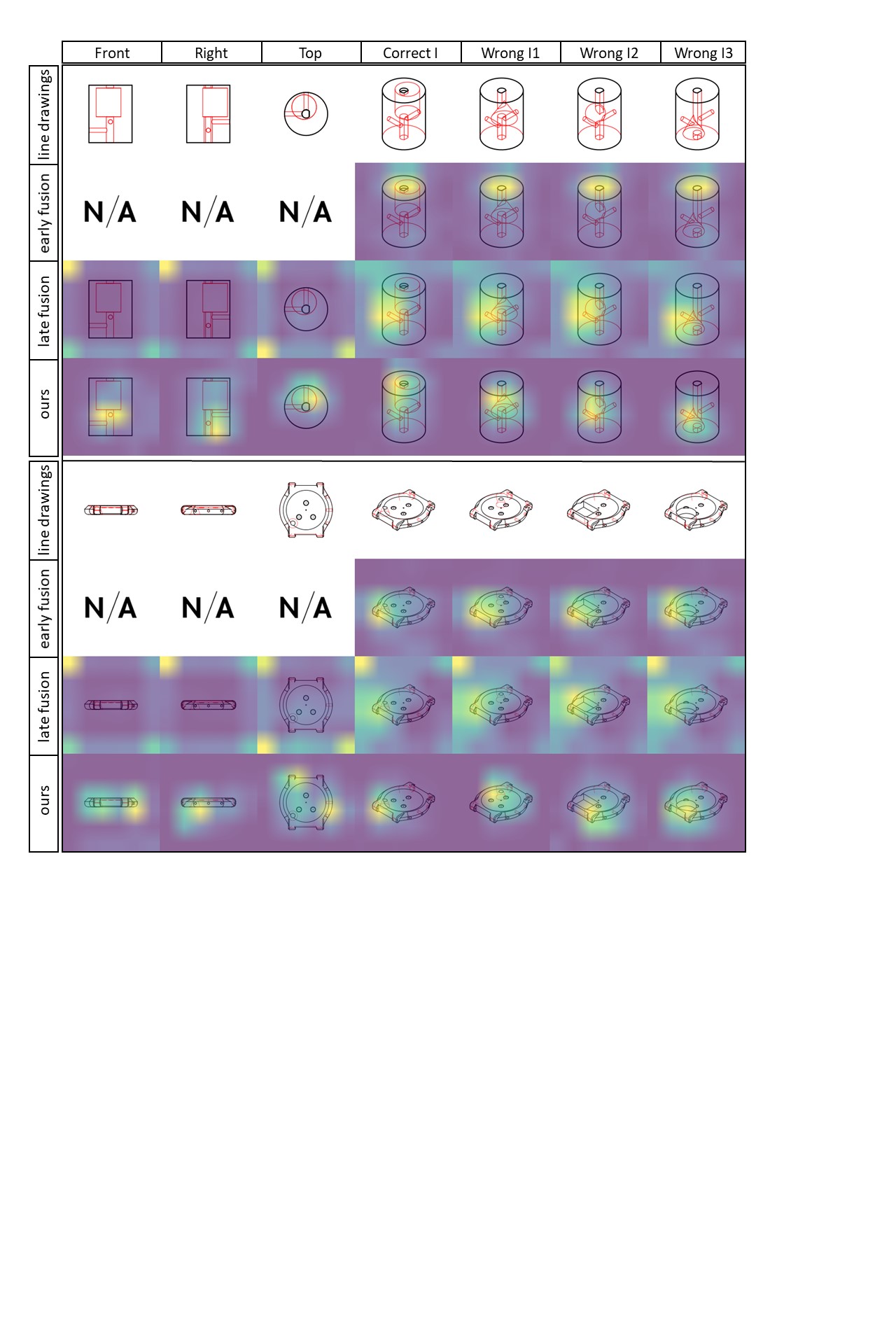}
% 	\vspace{-10mm}
\caption{\textbf{Attention maps for SL vs. SSL method in \textit{T2I} task.}}
\label{fig_supp_attn_maps_3}
\end{figure*}

\section{Additional Attention Maps for \textit{T2I}}
We provide more visualization results of attention maps to compare our method with supervised learning method, as in Figure~\ref{fig_supp_attn_maps_3}.
% We also provide some examples with wrong visualization results using our method, as in Figure\ref{fig_supp_attn_maps_4}.